\pdfoutput=1

\documentclass[11pt]{article}

\usepackage[final]{acl}

\usepackage{times}
\usepackage{latexsym}
\usepackage{times}
\usepackage{latexsym}
\usepackage{amsmath}
\usepackage{calrsfs}
\DeclareMathAlphabet{\pazocal}{OMS}{zplm}{m}{n}
\usepackage{booktabs}
\usepackage{graphicx}
\usepackage{tabularray}
\usepackage{multirow}
\usepackage{algorithm}
\usepackage{algpseudocode}
\usepackage{hyperref}
\usepackage[T1]{fontenc}

\usepackage[utf8]{inputenc}
\usepackage{textcomp}

\DeclareUnicodeCharacter{0101}{a} 
\DeclareUnicodeCharacter{010D}{c} 
\DeclareUnicodeCharacter{0113}{e} 
\DeclareUnicodeCharacter{014D}{o} 
\DeclareUnicodeCharacter{016B}{u} 
\DeclareUnicodeCharacter{1E25}{h} 
\DeclareUnicodeCharacter{0113}{e} 
\DeclareUnicodeCharacter{1E25}{h} 
\DeclareUnicodeCharacter{1E47}{n} 
\DeclareUnicodeCharacter{1E45}{n} 
\DeclareUnicodeCharacter{1E0D}{d} 
\DeclareUnicodeCharacter{0101}{a} 
\DeclareUnicodeCharacter{015B}{s} 
\DeclareUnicodeCharacter{014D}{o} 
\DeclareUnicodeCharacter{1E43}{m} 
\DeclareUnicodeCharacter{1E45}{n} 
\DeclareUnicodeCharacter{0113}{e} 
\DeclareUnicodeCharacter{1E25}{h} 
\DeclareUnicodeCharacter{1E47}{n} 
\DeclareUnicodeCharacter{1E45}{n} 
\DeclareUnicodeCharacter{1E0D}{d} 
\DeclareUnicodeCharacter{0101}{a} 
\DeclareUnicodeCharacter{015B}{s} 
\DeclareUnicodeCharacter{014D}{o} 
\DeclareUnicodeCharacter{1E43}{m} 
\usepackage{microtype}

\usepackage{inconsolata}

\title{Enhancing Low-Resource NMT with a Multilingual Encoder and Knowledge Distillation: A Case Study}

\title{Enhancing Low-Resource NMT with a Multilingual Encoder and Knowledge Distillation: A Case Study  }
\author{Aniruddha Roy\textsuperscript{1}, Pretam Ray\textsuperscript{1},~Ayush Maheshwari\textsuperscript{2},~Sudeshna Sarkar\textsuperscript{1},~Pawan Goyal\textsuperscript{1} \\
  \textsuperscript{1} Indian Institute of Technology Kharagpur ~~~  \textsuperscript{2} Vizzhy Inc, Bengaluru \\
  \texttt{\{aniruddha.roy, pretam.ray\}@iitkgp.ac.in, ayush.maheshwari@vizzhy.com,}\\
\texttt{\{sudeshna, pawang\}@cse.iitkgp.ac.in}
  }
\begin{document}

\maketitle
\begin{abstract}
Neural Machine Translation (NMT) remains a formidable challenge, especially when dealing with low-resource languages. Pre-trained sequence-to-sequence (seq2seq) multi-lingual models, such as mBART-50, have demonstrated impressive performance in various low-resource NMT tasks. However, their pre-training has been confined to 50 languages, leaving out support for numerous low-resource languages, particularly those spoken in the Indian subcontinent. Expanding mBART-50's language support requires complex pre-training, risking performance decline due to catastrophic forgetting. 
Considering these expanding challenges, this paper explores a framework that leverages the benefits of a pre-trained language model along with knowledge distillation in a seq2seq architecture to facilitate translation for low-resource languages, including those not covered by mBART-50. The proposed framework employs a multilingual encoder-based seq2seq model as the foundational architecture and subsequently uses complementary knowledge distillation techniques to mitigate the impact of imbalanced training. 
 Our framework is evaluated on three low-resource Indic languages in four Indic-to-Indic directions, yielding significant BLEU-4 and chrF improvements over baselines\footnote{Our code is publicly available at \url{https://github.com/raypretam/Two-step-low-res-NMT}}. Further, we conduct human evaluation to confirm effectiveness of our approach.
\end{abstract}

\section{Introduction}
Neural Machine Translation (NMT) models \cite{bahdanau2016neural, vaswani2017attention,liu-etal-2020-norm,khandelwal2021nearest} have shown impressive results on benchmark datasets, mainly containing large amounts of parallel data. However, these models face challenges when applied to low-resource languages or languages with rich and diverse morphology. Previous approaches have 
leveraged pre-trained models trained on extensive corpora \cite{weng2019acquiring,wang-etal-2022-understanding,liu2020multilingual,liu-etal-2021-copying,haddow-etal-2022-survey,10.1145/3555340,roy-etal-2022-meta} to address these limitations. 

Pre-trained multilingual seq2seq-based models based on an encoder-decoder framework such as mBART-50 \cite{liu2020multilingual} have been  successfully used for  various low-resource NMT tasks. Despite being pre-trained with 50 languages, it needs more support for numerous low-resource languages. Expanding the capabilities of mBART-50 to encompass new languages entails a cumbersome
process involving the collection of substantial amounts of monolingual data and the execution of pre-training with denoising objectives after initializing mBART-50. 
This process is time-consuming and may decrease performance on the initial 50 languages when incorporating new ones, a phenomenon known as catastrophic forgetting \cite{article}.

In contrast, encoder-based pretrained model XLM-R \cite{conneau2020unsupervised} is designed to accommodate 100 languages, making it suitable for a wide range of low-resource cross-lingual Natural Language Understanding (NLU) tasks. Both cross-lingual and Machine Translation (MT) functionalities share certain similarities. In cross-lingual scenarios, training and evaluation occur across different languages, while MT systems process input in one language and produce output in another. This distinction prompts several experimental research questions, including: 1) How does an XLM-R based NMT model perform on low-resource morphologically rich languages, particularly those not covered by mBART-50? 2) Given that low-resource NMT may be affected by training imbalances leading to performance degradation, can the application of knowledge distillation further enhance the results?

To address the two aforementioned experimental research questions, we utilize our base model, which follows a seq2seq framework. Here, we initialize the encoder with the multilingual pretrained model XLM-R large, while decoder layers are initialized from scratch, we call this base approach as XLM-MT. Similar frameworks have been explored in previous studies \cite{zhu2020incorporating,li2023multilingual}, with our approach sharing similarities with \cite{chen-etal-2022-towards}, who employed it for zero-shot cross-lingual NMT tasks and froze the embedding layers. However, our base approach differs in considering only decoder training. Thereafter, we apply complementary knowledge distillation (CKD) \cite{shao2022overcoming} to the base XLM-MT model to address training imbalances. The objective of this complementary knowledge distillation is to train the student model with knowledge which complements the teacher model and avoid knowledge forgetting, and we refer to this as XLM-MT+CKD. 
We empirically evaluate our model across three Indic languages and observe significant improvement in BLEU and chrF scores. Finally, we use human evaluation to assess the fluency, relatedness, and correctness of our output. Our contributions are as follows: \newline
    1. We repurpose the XLM-based seq2seq framework in conjunction with a complementary knowledge distillation approach to effectively design an NMT model for low-resource MT tasks. To the best of our knowledge, we are the first to integrate these two approaches effectively for NMT tasks.
    
\noindent    2.  We conduct comprehensive experiments on three Indian languages in four directions that are not included in mBART-50 and demonstrate the significance of our approach in enhancing translation results.

  \noindent  3. We also perform a detailed analysis of the results, including human evaluation and error analysis, for our proposed model.   
\section{Methodology}
Given a source language sentence $X = (x_1, x_2, \ldots, x_S)$, and its corresponding target language translation $Y = (y_1, y_2, \ldots, y_T)$, an NMT model is trained to predict the translated sequence $Y'$ using the maximum log-likelihood estimation (MLE) objective.  The probability of predicting the target sequence $Y'$ is computed as $p(Y'| X; \theta) = \prod_{t=1}^{T} p(y_t | y_{0:t-1}, x_{1:S}, \theta)$, where $\theta$ represents the model parameters.

\subsection{Base Model (XLM-MT)}
\label{sec:basemodel}
We initialize encoder layers and encoder embeddings with an unsupervised pre-trained multilingual model, XLM-R large \cite{conneau2020unsupervised} which is trained using masked language model objective. Then, we train the decoder from scratch while freezing the encoder parameters. During training, decoder parameters are learned with an MLE objective. The underlying assumption is that the pre-trained encoder parameters have already learnt a multilingual representation of the source language. As a result, only the decoder is trained using MLE objective while leveraging the encoder embeddings learned by the pre-trained model.
$ \pazocal{L}_{\theta_{dec}} = \sum_{(X,Y) \in D} \log P(Y|X;\theta_{dec})$
where $X$ and $Y$ represents the source  target sentence respectively from the dataset $D$. The parameter $\theta_{dec}$ refers to the parameters of the decoder layers and embedding.

\begin{algorithm}[!h]

\caption{Complementary Knowledge Distillation  }
\begin{algorithmic}[1]
\State \textbf{Input:} Training data $D$, the number of teachers $n$.
\State \textbf{Output:} Student model $S$.
\State Initialize $S$ and teacher models ($T_{1:n}$) with the base model, XLM-MT.
\While{not  converge}  
    \State randomly divides the training data $D$ in mutually exclusive  $n+1$ subsets $D_1, D_2, ... D_{n+1}$
    \For{\texttt{$t=1$} to \texttt{$n+1$}}
    \For{\texttt{$i=1$} to \texttt{$n$}}
      \State    Train $T_i$ on $D_{O(i,t)}$ 
        \EndFor
        \State Train $S$ on $D_{t}$ using Eq \ref{eq:eq1}
        \EndFor

    \For{\texttt{$i=1$} to \texttt{$n$}} 
    \State $T_i\gets S$ (At the end of
each epoch, reinitialize teacher models with the student model:)
    \EndFor
\EndWhile
\State \Return student model $S$
\end{algorithmic}\label{algo:1}
\end{algorithm}


\subsection{Complementary Knowledge Distillation}
\label{sec:kd}

Imbalances in training data lead to performance degradation in low-resource NMT due to catastrophic knowledge forgetting  \cite{lecun2002efficient,shao2022overcoming}.  We leverage complementary knowledge distillation (CKD) technique \cite{shao2022overcoming} to overcome this problem in low-resource MT. In CKD, $n$ teacher models and a student model $S$ are trained in a complementary manner such that $S$ learns from new training samples while teacher models dynamically provide complementary early samples knowledge to the $S$. In our case, both teacher and student models are initialized with the parameters of our base model, XLM-MT.

We divide the training set $D$ into $n+1$ mutually exclusive subsets for each epoch. The student model $S$ sequentially learns from $D_1$ to $D_{n+1}$ while the teacher models learn from all data splits except $D_t$. To determine the training data for the teacher models at timestep $t$, we utilize an ordering function, as shown in Eq \ref{eq:eq4} \cite{shao2022overcoming}. This ordering function covers all data splits except $D_t$, ensuring that the teacher models complement the student model.
\begin{align}\label{eq:eq4}
O(i,t) = \begin{Bmatrix}
i+t, & i+t\leq n+1 & \\ i+t-n-1, 
 & i+t> n+1
\end{Bmatrix}
\end{align}
where,  $i \in  \left \{ 1, 2, \ldots , n \right \} $ and \text{ }   $t \in  \left \{ 1, 2, \ldots, n + 1 \right \} $

In the process of word-level knowledge distillation, the student model $S$ benefits from an additional supervision signal, aligning its outputs with the probability outputs of the teacher model $T$. 
\begin{align}\label{eq:eq2}
 \pazocal{L}_{KD}({\theta}) 
 & = -\sum_{t=1}^T\sum_{k=1}^{|V|}\sum_{i=1}^n\frac{q_i(y_t=k|y_{<t},X)}{n} \nonumber
 \\ & \times \log\,p(y_t=k|y_{<t},X,\theta) 
\end{align}

where $|V|$ denotes the number of classes, $p$ denotes the prediction of student and $q_{i}$ is the prediction of teacher model $T_{i}$. To balance the distillation loss and the cross-entropy loss, we introduce a hyperparameter $\alpha$ for interpolation.
Finally, the overall objective function is 
\begin{align}\label{eq:eq1}
\pazocal{L}(\theta) = \alpha\cdot \pazocal{L}_{KD}(\theta) + (1 - \alpha)\cdot \pazocal{L}_{NLL}(\theta)
\end{align}
We employ a reinitialization technique \cite{8578552,lan2018knowledge} to facilitate two-way knowledge transfer. After each epoch, we reset the parameters of the teacher models using those of the student model. This reinitialization ensures that the student and teachers begin each epoch with identical settings.  We present the  training procedure for CKD in Algorithm \ref{algo:1}.
 We apply CKD to our base model in the following 2-step process. \\
\textbf{Step 1 - Initialization:} In this step, we initialize both the student and teacher models with the model obtained after the first step training ({\em c.f.}, Section \ref{sec:basemodel}).  This initialization ensures that the student model benefits from the knowledge acquired during the initial decoder training.\\
\textbf{Step 2 - CKD:} In this step, we apply the complementary KD technique ({\em c.f.},  Section \ref{sec:kd}) which enables the model to benefit from the transfer of complementary knowledge.
\section{Experimental Setup}\label{sec:im}


\noindent\textbf{Dataset:}
For our experiments, we specifically select three Indic languages, namely Kannada, and Punjabi that are not included in mBART-50, to assess the effectiveness of our approach. We use the Samanantar dataset~\cite{ramesh-etal-2022-samanantar}  for training all our NMT models which contains parallel sentences for 11 Indic language pairs. 
We consider three languages in 4 directions, namely Hindi-Kannada, Kannada-Hindi, Kannada-Punjabi, and Punjabi-Kannada, containing 2.1 million and 1.1 million parallel sentences respectively. 
We use the FLORES-200 \cite{nllb2022} containing 997 and 1012 sentences as our validation and test set respectively. 
\begin{table*}[ht]
\centering
\scalebox{0.95}{
\begin{tabular}{l|llll|llll}
\hline
\multirow{1}{*}{Model} 
                       & \multicolumn{1}{l}{hi-kn} & \multicolumn{1}{l}{kn-hi} & \multicolumn{1}{l}{kn-pa} & \multicolumn{1}{l}{pa-kn} & \multicolumn{1}{l}{hi-kn} & \multicolumn{1}{l}{kn-hi} & \multicolumn{1}{l}{kn-pa} & \multicolumn{1}{l}{pa-kn}  \\ \hline\hline
     ~ & ~ & \textbf{BLEU} & ~ & ~ & ~ &  \textbf{chrF} & ~ &  \\ \hline
     Transformer                  &   3.60  &   7.61    &  1.39    &  1.04  &  37.40   &  34.09     &  24.19    &  25.75     \\ \hline
   Sequence-KD \cite{kim2016sequencelevel}                 & 4.23   & 7.88    &  1.71    & 1.08  & 37.43 & 34.23 & 24.31 &25.89 \\\hline 
    
    mBERT-KD \cite{chen-etal-2020-distilling} & 4.73 & 8.67 &  2.01 & 1.31& 37.47 & 34.67 &  24.43 & 26.22 \\ \hline
    
    Selective KD \cite{wang2021selective} & 5.35 & 8.08 & 2.24 & 1.19 & 39.23& 35.02 &  24.57&26.78 \\ \hline 
    Transformer+CKD               &  4.51    &   8.89    &  3.23    &  1.98  &  38.54   &  35.13     &  24.54    &  27.01    \\ \hline

    mBERT-MT \cite{zhu2020incorporating} &4.98 & 10.23  & 3.78 & 4.17 &38.98&35.56&25.01&29.68 \\ \hline
    SixTp  \cite{chen-etal-2022-towards}             &  7.01  &   10.80      &   6.14    &   5.45    &  40.98  &   35.74      &    27.62   &  32.47\\ \hline
     XLM-MT (base)              &  6.08  &   8.75      &   6.01    &   2.98   &  40.38  &   35.38      &    27.12   &  28.11 \\ \hline

    XLM-MT + CKD  (ours)            &  \textbf{9.15}    &   \textbf{11.46}  &   \textbf{7.23}      &   \textbf{6.43}   & \textbf{41.11}   &   \textbf{35.88}   &    \textbf{29.12}     &  \textbf{33.98}   \\ \hline
   



  
\end{tabular}
}
\caption{Performance (BLEU-4 and chrF scores) of our model along with seven baseline models on the FLORES-200 dataset on 3 languages in 4 directions between Indic languages: Hindi (`hi'), Kannada (`kn'),  Punjabi (`pa'). We provide additional human evaluation results in Table \ref{tab:hum}.} 

\label{tab:tab1}
\end{table*}


  



\noindent\textbf{Implementation Details:}
We implement our approach using the Fairseq Toolkit~\cite{ott2019fairseq}. We use Adam optimizer~\cite{kingma2017adam} with $\beta_1=0.9$ and $\beta=0.98$. Following the work by \citet{chen-etal-2021-zero}, we use learning rates $5e-3$ and $1e-3$ for the base model and CKD, respectively. We set maximum updates of 200K for the base model training and 40K for the CKD. We use 12 layers with 16 attention heads in the decoder. We use  the `Large' variant of
XLM-R that has 550 million parameters for our experiments.  We set the number of teachers to 1 and $\alpha=0.95$. We set batch size $=32$k, and used beam size $=5$ throughout our experiments, and following \newcite{shao2022overcoming} we averaged the last five checkpoints. 
 We use BLEU-4 \cite{papineni-etal-2002-bleu} and chrF \cite{popovic-2015-chrf} score to evaluate our approach.
All the models have been trained on single A100 GPUs. None of the training methods consumed more than 96 hours.\\
\textbf{Baselines}
 We employ various baseline models for comparison with our approach. To ensure a fair assessment, we train all baseline models using identical training data and assess their performance on the Flores dataset.\\
\textbf{Transformer} \citep{vaswani2017attention}: We utilize a standard transformer-based encoder-decoder model, employing six layers for both the encoder and decoder.\\
\textbf{Word-level Knowledge Distillation} \citep{kim2016sequencelevel} is a conventional method applied to enhance NMT results by distilling knowledge at the word level. \\
\textbf{Sequence-level Knowledge Distillation } \citep{kim2016sequencelevel} is a conventional knowledge distillation technique applied to enhance NMT results by distilling knowledge at the sequence level.\\
\textbf{BERT-KD} \citep{chen-etal-2020-distilling} is a knowledge extracted from a fine-tuned BERT model is transferred to NMT models.
\\\textbf{Selective KD} \citep{wang2021selective} refers to the process of distilling and transferring specific, relevant knowledge from a teacher model to a student model. Instead of transferring all the knowledge indiscriminately, this approach involves selecting and distilling the most valuable and informative aspects of the teacher model's knowledge.\\ 
\textbf{mBERT-MT}\citep{zhu2020incorporating},  integrates BERT into the NMT process. Initially, BERT is employed to extract representations for an input sequence. Subsequently, these representations are fused with each layer of the NMT model's encoder and decoder using attention mechanisms.\\
\textbf{ sixTp} \citep{chen-etal-2022-towards} is a sequence-to-sequence (seq-to-seq) model. In its initialization, the encoders are initialized with the XLM-R large model, while the decoder is initialized randomly. The model undergoes a two-stage fine-tuning process. In the initial stage, the encoder layers are frozen, and fine-tuning is performed on the decoders. Subsequently, in the second stage, the model is trained in an end-to-end fashion.

\section{Results}

Table \ref{tab:tab1} presents the BLEU-4 and chrF results for Hindi to Kannada, Kannada to Punjabi in both directions. It is noteworthy that Hindi, and Punjabi belong to the Indo-Aryan language family, while Kannada belongs to the Dravidian family. 
We compare our results against seven competitive baselines, namely, vanilla transformer, knowledge distillation techniques, transformer with CKD, two step training techniques using mBERT and SixTp, which is XLM-R based model. 
We observe that  XLM-MT + CKD achieves BLEU scores within the range of (6.43 to 11.46) consistently surpassing the baselines. We observe an average improvements of 1.22-5.15 BLEU scores across all language pairs.
We also present chrF scores in Table \ref{tab:tab1}. Notably, XLM-MT+CKD consistently demonstrates its superiority, outperforming all the baselines with averages of  0.82-4.66 chrF score, across all language pairs. 
  Further, we conduct human evaluation to assess the fluency,  relatedness and correctness of the generated text. We present human evaluation results of sixTp and our model, XLM-MT+CKD in Table \ref{tab:hum}. 
  
We also investigate various variants of our model to validate the effectiveness of our architecture and present results in Table \ref{tab:tab3}. Additionally, we conduct comprehensive error analysis in Section \ref{sec:case_study}.    
\begin{table*}[]
    \centering
    \begin{tabular}{lllllll}
    \hline
        Model & hi-kn & kn-hi & kn-pa & pa-kn  \\ \hline \hline
        Transformer  &   3.60  &   7.61    &  2.39    &  1.04    \\ \hline
        $\text{Enc}^{\text{XLM-R}}_{\text{train}}$ + Dec & 4.72& 8.32 & 5.75& 2.11 \\ \hline
        $\text{Enc}^{\text{XLM-R}}_{\text{no-train}}$ + Dec & 6.08& 8.75& 6.01 & 2.98 \\ \hline
          SixTp              &  7.01  &   10.80      &   6.14    &   5.45  \\  \hline
        XLM-MT + CKD              &  \textbf{9.15}    &   \textbf{11.46}  &   \textbf{7.23}      &   \textbf{6.43}   \\  \hline\hline
    \end{tabular}
    \caption{Performances (BLEU-4 scores) of our model along with its variants. The score in \textbf{bold} shows the best scores for the corresponding language pair. Enc + Dec refers to the transformer model without XLM initialization. $\text{Enc}^{\text{XLM-R}}_{\text{train}}$ + Dec refers to joint training of XLM-R based encoder and decoder. $\text{Enc}^{\text{XLM-R}}_{\text{no-train}} + $ + Dec refers to only decoder training.}
    \label{tab:tab3}
\end{table*}

\section{Analysis}

\noindent\textbf{How would the method perform with the languages that mBART-50 supported?} 
In addition to the language pairs outlined in Section \ref{sec:im}, we extend our exploration to include language pairs supported by mBART-50, facilitating effective comparisons with the mBART-50 model. We extracted three language pairs from the Samantar dataset—namely, Hindi-Bengali, Telugu-Hindi, and Marathi-Hindi—and compared our approach with mBART-50. We present the results in Table \ref{tab:mbert}. mBART-50 achieves BLEU-4 scores of 9.35 and 17.06 for the language pairs of Hindi-Bengali and Marathi-Hindi respectively, surpassing the performance of XLM-MT+CKD model. For the Hindi-Telugu pair, our model XLM-MT+CKD achieves better performance than mBART-50.  

\begin{table}[]
    \centering
    \begin{tabular}{lllllll}
    \hline
        Model & hi-bn & mr-hi & hi-te  \\ \hline \hline
        mBART-50 & \textbf{9.25} &  \textbf{17.06} & 9.35 \\ \hline
        XLM-MT  & 8.13  & 15.78  & 11.56 \\  \hline
        XLM-MT + CKD  & 8.67  & 15.81 & \textbf{12.01}  \\  \hline\hline
    \end{tabular}
    \caption{Performances BLEU-4  of our model along with mBART-50 model on the FLORES-200 dataset for the translation between Indic languages: Hindi (`hi'), Telugu (`te'), and Bengali (`bn'). }
    \label{tab:mbert}
\end{table}
\subsection{Analysis of different model variants}

The aim of this analysis is to assess the effectiveness of our model with different approaches in addressing the challenges of machine translation, particularly for low-resource and morphologically rich languages. 
The obtained BLEU scores are presented in Table \ref{tab:tab3}.

\noindent\textbf{Enc + Dec:} To assess the importance of pre-training initialization in the encoder, we compare the performance of XLM-MT, which is initialized with XLM-R large, against a randomly initialized model. We observe that the encoder initialized with XLM-R large produces better performance than the randomly initialized encoder.

\noindent\textbf{$\text{Enc}^{\text{XLM-R}}_{\text{train}}$ + Dec,  $\text{Enc}^{\text{XLM-R}}_{\text{no-train}}$ + Dec:} To analyze the effectiveness of the two-stage training process employed in XLM-MT, we experiment with two different settings: (a) training encoder and decoder jointly (the second stage), denoted as $\text{Enc}^{\text{XLM-R}}_{\text{train}}$ + Dec, and (b)  only training the decoder (the first stage), denoted as $\text{Enc}^{\text{XLM-R}}_{\text{no-train}}$ + Dec. From Table \ref{tab:tab3}, we clearly see the effectiveness of two-stage training compared to only one of these stages across all language pairs. \\


\begin{table*}[]
\centering
\begin{tabular}{lllll}
\hline\hline
\multicolumn{1}{l|}{Metric}      & \multicolumn{1}{l|}{hi-kn} & \multicolumn{1}{l|}{kn-hi} & \multicolumn{1}{l|}{kn-pa} & \multicolumn{1}{l}{pa-kn} \\ \hline\hline
\multicolumn{5}{c}{SixTp}                                                                                                                                                                                                                          \\ \hline
\multicolumn{1}{l|}{Fluency}     & \multicolumn{1}{l|}{3.13}      & \multicolumn{1}{l|}{3.71}      & \multicolumn{1}{l|}{2.47}      & \multicolumn{1}{l}{1.97}  \\ \hline
\multicolumn{1}{l|}{Correctness} & \multicolumn{1}{l|}{3.04}      & \multicolumn{1}{l|}{3.68}      & \multicolumn{1}{l|}{2.40}      & \multicolumn{1}{l}{\textbf{1.87}}  \\ \hline
\multicolumn{1}{l|}{Relatedness} & \multicolumn{1}{l|}{3.18}      & \multicolumn{1}{l|}{3.75}      & \multicolumn{1}{l|}{2.33}      & \multicolumn{1}{l}{1.92}  \\ \hline
\multicolumn{5}{c}{XLM-MT + CKD}                                                                                                                                                                                                                         \\ \hline
\multicolumn{1}{l|}{Fluency}     & \multicolumn{1}{l|}{\textbf{3.23}}      & \multicolumn{1}{l|}{\textbf{3.75}}      & \multicolumn{1}{l|}{\textbf{2.53}}      & \multicolumn{1}{l}{\textbf{2.01}}     \\ \hline
\multicolumn{1}{l|}{Correctness} & \multicolumn{1}{l|}{\textbf{3.71}}      & \multicolumn{1}{l|}{\textbf{3.92}}      & \multicolumn{1}{l|}{\textbf{2.47}}      & \multicolumn{1}{l}{1.85}    \\ \hline
\multicolumn{1}{l|}{Relatedness} & \multicolumn{1}{l|}{\textbf{3.43}}      & \multicolumn{1}{l|}{\textbf{3.78}}      & \multicolumn{1}{l|}{\textbf{2.51}}      & \multicolumn{1}{l}{\textbf{1.97}}  \\ \hline\hline
\end{tabular}

\caption{Human evaluation results of our approach sixTp and XLM-MT+ CKD for three languages in four directions.
The three metrics are Fluency, Relatedness, and Correctness, respectively.}

\label{tab:hum}
\end{table*}

\section{Human Evaluation}\label{sec:human-evaluation}
We follow a procedure similar to previous studies \cite{chi2019crosslingual,maurya2021zmbart} to assess the quality of translated sentences in three Indic languages in four Indic-to-Indic language pairs. We randomly selected 50 test data-points for each language pair for evaluation. Three key metrics are used to evaluate the translated sentences: fluency, relatedness, and correctness. Fluency refers to the smoothness and coherence of the generated text, evaluating how well the sentences flow and adhere to grammatical rules. Relatedness measures how well the translated sentences are connected to the given ground truth sentences and capture its key information. Correctness assesses the accuracy and appropriateness of the translated sentences in terms of their meaning and semantics. 
We present the translated sentences (randomly shuffled) from two models XLM-MT and XLM-MT+CKD to three language experts for each language pair. The selected 12 experts are well versed in the corresponding target language including English. The experts attained a minimum of graduate degree in English and have native proficiency in the target language. The experts are informed about the task and  were  renumerated as per industry standard norms. The experts rated the sentences on a 5-point scale, with 1 indicating very bad and 5 indicating very good, for each of the three metrics. 
The final numbers are in Table \ref{tab:hum}.
These are calculated by averaging all the experts’ responses for each parameter. The annotation experts received compensation according to industry standards for their work. We briefed them on the objectives and explicit usage of their annotations
\begin{table*}[!ht]
    \centering
    \begin{tabular}{p{0.25\linewidth} p{0.75\linewidth}}
    \hline
        \textbf{1. Source (Kannada)}: & Udaharanage, obbaru, motaru karugale rastegala abhivrd'dhige mula karana endu helabahudu. \\
        ~   &   \textbf{Translation:} {\color{orange}For example, one could say that motor cars were the root cause of the development of roads.} \\
        \textbf{Reference (Punjabi):} & 
         Udaharana vajon, koi kahi sakada hai ki motara kara sarakan nu zaruri taura'te vikas vala lai jandi hai.\\    
         ~ &    \textbf{Translation:}  {\color{orange}For example, one could say that the motor car essentially leads to development of roads.} \\
        \textbf{XLM-MT+CKD:} &  Udaharana vajon, koi kahi sakada hai ki motara kara sarakan nu zaruri taura'te vikas vala lai jandi hai.\\
         ~ & \textbf{Translation:}  {\color{orange}For example, one could say that the motor car essentially leads to development of roads.} \\ \hline \hline
        \textbf{2. Source (Hindi) :} & kuchh any visheshagyon kee tarah, unhen is baat par sandeh hai ki kya madhumeh ko theek kiya ja sakata hai, yah dekhate hue ki in nishkarshon kee un logon ke lie koee praasangikata nahin hai jinhen pahale se hee taip 1 madhumeh hai. \\
        ~   &   \textbf{Translation:} {\color{orange}Like some other experts, he is skeptical about whether diabetes can be cured, noting that these findings have no relevance to people who already have type 1 diabetes.} \\
        \textbf{Reference (Kannada):} & Madhumēhavannu guṇapaḍisalu sādhyavē emba bagge itare itara kelavu tajñarante avaru kūḍa sanśaya vyaktapaḍisuttāre, ī sanśōdhanegaḷu īgāgalē ṭaip 1 madhumēha hondiruva janarige yāvudē prayōjanagaḷannu nīḍilla.\\
        ~   &   \textbf{Translation:} {\color{orange}Like some other experts, he doubts whether diabetes can be cured, because these conclusions are not practical for people who have previously had type 1 diabetes.} \\
        \textbf{XLM-MT+CKD:} & Itara kelavu tajñarante, avaru madhumēhavannu guṇapaḍisabahudē endu sanśayapaḍuttāre, ēkendare ī tīrmānagaḷu ī hinde ṭaip 1 madhumēha hondiruva vyaktigaḷige prāyōgikavāgiruvudilla.\\
        ~   &   \textbf{Translation:} {\color{orange}Like some other experts, he doubts whether diabetes can be cured, because these conclusions are not practical for people who have previously had type 1 diabetes.} \\ \hline \hline
    \end{tabular}
    \caption{Sample outputs generated from our proposed approach, where the target languages' source language and translations are specified for each reference.}
    \label{tab:case}
\end{table*}

\section{Case Study}\label{sec:case_study}
Table \ref{tab:case} presents several example sentences and their translations by our proposed approach. Notably, there are specific issues with reference 1 in the XLM-MT translations. In reference 1, XLM-MT incorrectly translates the sentence using a wrong gender concept, whereas XLM-MT+CKD translates correctly.
Regarding the Kannada sentence in reference 2, the XLM-MT and XLM-MT+CKD approaches provide a correct and meaningful translation, albeit with some paraphrasing.

\section{Related Work}Neural Machine Translation (NMT) aims to translate a given source sentence into a target sentence. Typically, an NMT model comprises an encoder, a decoder, and an attention mechanism. The encoder transforms the input sequence into hidden representations while the decoder maps these representations to the target sequence. The attention mechanism, pioneered by \cite{bahdanau2016neural}, enhances alignment between words in the source and target languages. 
Different architectures can be employed for the encoder and decoder, including LSTM (Long Short-Term Memory), CNN (Convolutional Neural Network), and Transformer. The Transformer architecture, introduced by \cite{vaswani2017attention}, consists of three sublayers. Transformer has demonstrated state-of-the-art performance in NMT tasks \cite{barrault-etal-2019-findings}.

Prior studies \cite{imamura-sumita-2019-recycling,lample2019crosslingual,yang2022making,weng2019acquiring,ma2020xlmt,zhu2020incorporating} have investigated the integration of pre-trained language encoders into NMT models to bolster supervised translation performance. \cite{zhu2020incorporating} introduce a BERT-fused model that extracts representations from input sentences and integrates them into the encoder and decoder using attention mechanisms. Recent research \cite{song2019mass}  focuses on developing and refining encoder-decoder-based multilingual trained language models for NMT.  \cite{liu2020multilingualdenoisingpretrainingneural} present mBART, a Transformer-based encoder-decoder model explicitly tailored for NMT applications. Wei et al. finetune the multilingual encoder-based model for low-resource NMT, and they focus on improving the MPE for a more
universal representation across languages. \cite{chen-etal-2021-zero,chen-etal-2022-towards} have examined a two-stage framework utilizing an encoder-based multilingual language model for zero-shot neural machine translation.

Numerous studies in NMT have incorporated the Knowledge Distillation (KD) framework. \cite{kim2016sequencelevel} introduced word-level KD for NMT and later proposed sequence-level KD to enhance overall performance. Investigating the efficacy of various token types in KD, \cite{wang2021selective} suggested strategies for selective KD. \cite{wu2020skip} successfully transferred internal hidden states from teacher models to students, achieving positive results. Various KD approaches have also been employed in non-auto-regressive Machine Translation tasks to enhance outcomes. \cite{gu2018nonautoregressive} improved non-autoregressive model performance by distilling information from an autoregressive model. \cite{zhou2021understanding} conducted systematic experiments highlighting the importance of knowledge distillation in training non-auto-regressive models, showing its ability to reduce dataset complexity and help model variations in output data. In the realm of multilingual NMT, \cite{baziotis-etal-2020-language} used language models as instructors for low-resource NMT models. \cite{chen-etal-2020-distilling} extracted knowledge from fine-tuned BERT and transferred it to NMT models. Furthermore, \cite{feng-etal-2021-guiding} and \cite{zhou2021understanding} employed KD to introduce forward-looking information into the teacher-forcing training of NMT models.

   

\section{Conclusion}
In this paper, we empirically explored the methods for improving low-resource NMT, particularly for Indic languages. We investigated several strategies for initialization of encoder and decoder, along with the knowledge distillation techniques.
We conducted experiment on three low-resource Indic languages in four Indic-to-Indic directions belonging to two language families, specifically focusing on those not covered by mBART-50. Further, we perform additional analysis on languages supported by mBART-50 and high-resource language pairs.

\section*{Limitations}
A limitation of this study is the increased training time required for the XLM-MT+CKD model due to its addition of complementary knowledge distillation. Furthermore, our validation is limited to low-resource machine translation tasks, although seq2seq models
have the potential to be utilized for a wide range
of generation tasks, including Question Generation and Summarization in both monolingual and
cross-lingual contexts. 

\section*{Acknowledgement}
This work was supported in part by the National Language Translation Mission (NLTM): Bhashini project by the Government of India.

\bibliography{anthology,custom}

\begin{thebibliography}{44}
\expandafter\ifx\csname natexlab\endcsname\relax\def\natexlab#1{#1}\fi

\bibitem[{Bahdanau et~al.(2016)Bahdanau, Cho, and Bengio}]{bahdanau2016neural}
Dzmitry Bahdanau, Kyunghyun Cho, and Yoshua Bengio. 2016.
\newblock \href {http://arxiv.org/abs/1409.0473} {Neural machine translation by jointly learning to align and translate}.

\bibitem[{Barrault et~al.(2019)Barrault, Bojar, Costa-juss{\`a}, Federmann, Fishel, Graham, Haddow, Huck, Koehn, Malmasi, Monz, M{\"u}ller, Pal, Post, and Zampieri}]{barrault-etal-2019-findings}
Lo{\"\i}c Barrault, Ond{\v{r}}ej Bojar, Marta~R. Costa-juss{\`a}, Christian Federmann, Mark Fishel, Yvette Graham, Barry Haddow, Matthias Huck, Philipp Koehn, Shervin Malmasi, Christof Monz, Mathias M{\"u}ller, Santanu Pal, Matt Post, and Marcos Zampieri. 2019.
\newblock \href {https://doi.org/10.18653/v1/W19-5301} {Findings of the 2019 conference on machine translation ({WMT}19)}.
\newblock In \emph{Proceedings of the Fourth Conference on Machine Translation (Volume 2: Shared Task Papers, Day 1)}, pages 1--61, Florence, Italy. Association for Computational Linguistics.

\bibitem[{Baziotis et~al.(2020)Baziotis, Haddow, and Birch}]{baziotis-etal-2020-language}
Christos Baziotis, Barry Haddow, and Alexandra Birch. 2020.
\newblock \href {https://doi.org/10.18653/v1/2020.emnlp-main.615} {Language model prior for low-resource neural machine translation}.
\newblock In \emph{Proceedings of the 2020 Conference on Empirical Methods in Natural Language Processing (EMNLP)}, pages 7622--7634, Online. Association for Computational Linguistics.

\bibitem[{Chen et~al.(2021)Chen, Ma, Chen, Dong, Zhang, Pan, Wang, and Wei}]{chen-etal-2021-zero}
Guanhua Chen, Shuming Ma, Yun Chen, Li~Dong, Dongdong Zhang, Jia Pan, Wenping Wang, and Furu Wei. 2021.
\newblock \href {https://doi.org/10.18653/v1/2021.emnlp-main.2} {Zero-shot cross-lingual transfer of neural machine translation with multilingual pretrained encoders}.
\newblock In \emph{Proceedings of the 2021 Conference on Empirical Methods in Natural Language Processing}, pages 15--26, Online and Punta Cana, Dominican Republic. Association for Computational Linguistics.

\bibitem[{Chen et~al.(2022)Chen, Ma, Chen, Zhang, Pan, Wang, and Wei}]{chen-etal-2022-towards}
Guanhua Chen, Shuming Ma, Yun Chen, Dongdong Zhang, Jia Pan, Wenping Wang, and Furu Wei. 2022.
\newblock \href {https://doi.org/10.18653/v1/2022.acl-long.12} {Towards making the most of cross-lingual transfer for zero-shot neural machine translation}.
\newblock In \emph{Proceedings of the 60th Annual Meeting of the Association for Computational Linguistics (Volume 1: Long Papers)}, pages 142--157, Dublin, Ireland. Association for Computational Linguistics.

\bibitem[{Chen et~al.(2020)Chen, Gan, Cheng, Liu, and Liu}]{chen-etal-2020-distilling}
Yen-Chun Chen, Zhe Gan, Yu~Cheng, Jingzhou Liu, and Jingjing Liu. 2020.
\newblock \href {https://doi.org/10.18653/v1/2020.acl-main.705} {Distilling knowledge learned in {BERT} for text generation}.
\newblock In \emph{Proceedings of the 58th Annual Meeting of the Association for Computational Linguistics}, pages 7893--7905, Online. Association for Computational Linguistics.

\bibitem[{Chi et~al.(2019)Chi, Dong, Wei, Wang, Mao, and Huang}]{chi2019crosslingual}
Zewen Chi, Li~Dong, Furu Wei, Wenhui Wang, Xian-Ling Mao, and Heyan Huang. 2019.
\newblock \href {http://arxiv.org/abs/1909.10481} {Cross-lingual natural language generation via pre-training}.

\bibitem[{Conneau et~al.(2020)Conneau, Khandelwal, Goyal, Chaudhary, Wenzek, Guzmán, Grave, Ott, Zettlemoyer, and Stoyanov}]{conneau2020unsupervised}
Alexis Conneau, Kartikay Khandelwal, Naman Goyal, Vishrav Chaudhary, Guillaume Wenzek, Francisco Guzmán, Edouard Grave, Myle Ott, Luke Zettlemoyer, and Veselin Stoyanov. 2020.
\newblock \href {http://arxiv.org/abs/1911.02116} {Unsupervised cross-lingual representation learning at scale}.

\bibitem[{Conneau and Lample(2019)}]{lample2019crosslingual}
Alexis Conneau and Guillaume Lample. 2019.
\newblock Cross-lingual language model pretraining.
\newblock \emph{Advances in neural information processing systems}, 32.

\bibitem[{Feng et~al.(2021)Feng, Gu, Guo, Yang, and Shao}]{feng-etal-2021-guiding}
Yang Feng, Shuhao Gu, Dengji Guo, Zhengxin Yang, and Chenze Shao. 2021.
\newblock \href {https://doi.org/10.18653/v1/2021.acl-long.223} {Guiding teacher forcing with seer forcing for neural machine translation}.
\newblock In \emph{Proceedings of the 59th Annual Meeting of the Association for Computational Linguistics and the 11th International Joint Conference on Natural Language Processing (Volume 1: Long Papers)}, pages 2862--2872, Online. Association for Computational Linguistics.

\bibitem[{French(1999)}]{article}
Robert French. 1999.
\newblock \href {https://doi.org/10.1016/S1364-6613(99)01294-2} {Catastrophic forgetting in connectionist networks}.
\newblock \emph{Trends in cognitive sciences}, 3:128--135.

\bibitem[{Gu et~al.(2018)Gu, Bradbury, Xiong, Li, and Socher}]{gu2018nonautoregressive}
J~Gu, J~Bradbury, C~Xiong, VOK Li, and R~Socher. 2018.
\newblock Non-autoregressive neural machine translation.
\newblock In \emph{International Conference on Learning Representations (ICLR)}.

\bibitem[{Haddow et~al.(2022)Haddow, Bawden, Miceli~Barone, Helcl, and Birch}]{haddow-etal-2022-survey}
Barry Haddow, Rachel Bawden, Antonio~Valerio Miceli~Barone, Jind{\v{r}}ich Helcl, and Alexandra Birch. 2022.
\newblock \href {https://doi.org/10.1162/coli_a_00446} {Survey of low-resource machine translation}.
\newblock \emph{Computational Linguistics}, 48(3):673--732.

\bibitem[{Imamura and Sumita(2019)}]{imamura-sumita-2019-recycling}
Kenji Imamura and Eiichiro Sumita. 2019.
\newblock \href {https://doi.org/10.18653/v1/D19-5603} {Recycling a pre-trained {BERT} encoder for neural machine translation}.
\newblock In \emph{Proceedings of the 3rd Workshop on Neural Generation and Translation}, pages 23--31, Hong Kong. Association for Computational Linguistics.

\bibitem[{Khandelwal et~al.()Khandelwal, Fan, Jurafsky, Zettlemoyer, and Lewis}]{khandelwal2021nearest}
Urvashi Khandelwal, Angela Fan, Dan Jurafsky, Luke Zettlemoyer, and Mike Lewis.
\newblock Nearest neighbor machine translation.
\newblock In \emph{International Conference on Learning Representations}.

\bibitem[{Kim and Rush(2016)}]{kim2016sequencelevel}
Yoon Kim and Alexander~M. Rush. 2016.
\newblock \href {http://arxiv.org/abs/1606.07947} {Sequence-level knowledge distillation}.

\bibitem[{Kingma and Ba(2017)}]{kingma2017adam}
Diederik~P. Kingma and Jimmy Ba. 2017.
\newblock \href {http://arxiv.org/abs/1412.6980} {Adam: A method for stochastic optimization}.

\bibitem[{LeCun et~al.(2002)LeCun, Bottou, Orr, and M{\"u}ller}]{lecun2002efficient}
Yann LeCun, L{\'e}on Bottou, Genevieve~B Orr, and Klaus-Robert M{\"u}ller. 2002.
\newblock Efficient backprop.
\newblock In \emph{Neural networks: Tricks of the trade}, pages 9--50. Springer.

\bibitem[{Li et~al.(2023)Li, Rasooli, Patel, and Callison-Burch}]{li2023multilingual}
Bryan Li, Mohammad~Sadegh Rasooli, Ajay Patel, and Chris Callison-Burch. 2023.
\newblock Multilingual bidirectional unsupervised translation through multilingual finetuning and back-translation.
\newblock In \emph{Proceedings of the Sixth Workshop on Technologies for Machine Translation of Low-Resource Languages (LoResMT 2023)}, pages 16--31.

\bibitem[{Liu et~al.(2020{\natexlab{a}})Liu, Lai, Wong, and Chao}]{liu-etal-2020-norm}
Xuebo Liu, Houtim Lai, Derek~F. Wong, and Lidia~S. Chao. 2020{\natexlab{a}}.
\newblock \href {https://doi.org/10.18653/v1/2020.acl-main.41} {Norm-based curriculum learning for neural machine translation}.
\newblock In \emph{Proceedings of the 58th Annual Meeting of the Association for Computational Linguistics}, pages 427--436, Online. Association for Computational Linguistics.

\bibitem[{Liu et~al.(2021)Liu, Wang, Wong, Ding, Chao, Shi, and Tu}]{liu-etal-2021-copying}
Xuebo Liu, Longyue Wang, Derek~F. Wong, Liang Ding, Lidia~S. Chao, Shuming Shi, and Zhaopeng Tu. 2021.
\newblock \href {https://doi.org/10.18653/v1/2021.findings-acl.373} {On the copying behaviors of pre-training for neural machine translation}.
\newblock In \emph{Findings of the Association for Computational Linguistics: ACL-IJCNLP 2021}, pages 4265--4275, Online. Association for Computational Linguistics.

\bibitem[{Liu et~al.(2020{\natexlab{b}})Liu, Gu, Goyal, Li, Edunov, Ghazvininejad, Lewis, and Zettlemoyer}]{liu2020multilingual}
Yinhan Liu, Jiatao Gu, Naman Goyal, Xian Li, Sergey Edunov, Marjan Ghazvininejad, Mike Lewis, and Luke Zettlemoyer. 2020{\natexlab{b}}.
\newblock \href {http://arxiv.org/abs/2001.08210} {Multilingual denoising pre-training for neural machine translation}.

\bibitem[{Liu et~al.(2020{\natexlab{c}})Liu, Gu, Goyal, Li, Edunov, Ghazvininejad, Lewis, and Zettlemoyer}]{liu2020multilingualdenoisingpretrainingneural}
Yinhan Liu, Jiatao Gu, Naman Goyal, Xian Li, Sergey Edunov, Marjan Ghazvininejad, Mike Lewis, and Luke Zettlemoyer. 2020{\natexlab{c}}.
\newblock \href {http://arxiv.org/abs/2001.08210} {Multilingual denoising pre-training for neural machine translation}.

\bibitem[{Ma et~al.(2020)Ma, Yang, Huang, Chi, Dong, Zhang, Awadalla, Muzio, Eriguchi, Singhal, Song, Menezes, and Wei}]{ma2020xlmt}
Shuming Ma, Jian Yang, Haoyang Huang, Zewen Chi, Li~Dong, Dongdong Zhang, Hany~Hassan Awadalla, Alexandre Muzio, Akiko Eriguchi, Saksham Singhal, Xia Song, Arul Menezes, and Furu Wei. 2020.
\newblock \href {http://arxiv.org/abs/2012.15547} {Xlm-t: Scaling up multilingual machine translation with pretrained cross-lingual transformer encoders}.

\bibitem[{Maurya et~al.(2021)Maurya, Desarkar, Kano, and Deepshikha}]{maurya2021zmbart}
Kaushal~Kumar Maurya, Maunendra~Sankar Desarkar, Yoshinobu Kano, and Kumari Deepshikha. 2021.
\newblock \href {http://arxiv.org/abs/2106.01597} {Zmbart: An unsupervised cross-lingual transfer framework for language generation}.

\bibitem[{Ott et~al.(2019)Ott, Edunov, Baevski, Fan, Gross, Ng, Grangier, and Auli}]{ott2019fairseq}
Myle Ott, Sergey Edunov, Alexei Baevski, Angela Fan, Sam Gross, Nathan Ng, David Grangier, and Michael Auli. 2019.
\newblock fairseq: A fast, extensible toolkit for sequence modeling.
\newblock In \emph{Proceedings of NAACL-HLT 2019: Demonstrations}.

\bibitem[{Papineni et~al.(2002)Papineni, Roukos, Ward, and Zhu}]{papineni-etal-2002-bleu}
Kishore Papineni, Salim Roukos, Todd Ward, and Wei-Jing Zhu. 2002.
\newblock \href {https://doi.org/10.3115/1073083.1073135} {{B}leu: a method for automatic evaluation of machine translation}.
\newblock In \emph{Proceedings of the 40th Annual Meeting of the Association for Computational Linguistics}, pages 311--318, Philadelphia, Pennsylvania, USA. Association for Computational Linguistics.

\bibitem[{Popovi{\'c}(2015)}]{popovic-2015-chrf}
Maja Popovi{\'c}. 2015.
\newblock \href {https://doi.org/10.18653/v1/W15-3049} {chr{F}: character n-gram {F}-score for automatic {MT} evaluation}.
\newblock In \emph{Proceedings of the Tenth Workshop on Statistical Machine Translation}, pages 392--395, Lisbon, Portugal. Association for Computational Linguistics.

\bibitem[{Ramesh et~al.(2022)Ramesh, Doddapaneni, Bheemaraj, Jobanputra, AK, Sharma, Sahoo, Diddee, J, Kakwani, Kumar, Pradeep, Nagaraj, Deepak, Raghavan, Kunchukuttan, Kumar, and Khapra}]{ramesh-etal-2022-samanantar}
Gowtham Ramesh, Sumanth Doddapaneni, Aravinth Bheemaraj, Mayank Jobanputra, Raghavan AK, Ajitesh Sharma, Sujit Sahoo, Harshita Diddee, Mahalakshmi J, Divyanshu Kakwani, Navneet Kumar, Aswin Pradeep, Srihari Nagaraj, Kumar Deepak, Vivek Raghavan, Anoop Kunchukuttan, Pratyush Kumar, and Mitesh~Shantadevi Khapra. 2022.
\newblock \href {https://doi.org/10.1162/tacl_a_00452} {Samanantar: The largest publicly available parallel corpora collection for 11 {I}ndic languages}.
\newblock \emph{Transactions of the Association for Computational Linguistics}, 10:145--162.

\bibitem[{Roy et~al.(2023)Roy, Sharma, Sarkar, and Goyal}]{10.1145/3555340}
Aniruddha Roy, Isha Sharma, Sudeshna Sarkar, and Pawan Goyal. 2023.
\newblock \href {https://doi.org/10.1145/3555340} {Meta-ed: Cross-lingual event detection using meta-learning for indian languages}.
\newblock \emph{ACM Trans. Asian Low-Resour. Lang. Inf. Process.}, 22(2).

\bibitem[{Roy et~al.(2022)Roy, Thakur, Sharma, Gupta, Krishna, Sarkar, and Goyal}]{roy-etal-2022-meta}
Aniruddha Roy, Rupak~Kumar Thakur, Isha Sharma, Ashim Gupta, Amrith Krishna, Sudeshna Sarkar, and Pawan Goyal. 2022.
\newblock \href {https://aclanthology.org/2022.coling-1.373} {Does meta-learning help m{BERT} for few-shot question generation in a cross-lingual transfer setting for indic languages?}
\newblock In \emph{Proceedings of the 29th International Conference on Computational Linguistics}, pages 4251--4257, Gyeongju, Republic of Korea. International Committee on Computational Linguistics.

\bibitem[{Shao and Feng(2022)}]{shao2022overcoming}
Chenze Shao and Yang Feng. 2022.
\newblock Overcoming catastrophic forgetting beyond continual learning: Balanced training for neural machine translation.
\newblock In \emph{Proceedings of the 60th Annual Meeting of the Association for Computational Linguistics (Volume 1: Long Papers)}, pages 2023--2036.

\bibitem[{Song et~al.(2019)Song, Tan, Qin, Lu, and Liu}]{song2019mass}
Kaitao Song, Xu~Tan, Tao Qin, Jianfeng Lu, and Tie-Yan Liu. 2019.
\newblock \href {http://arxiv.org/abs/1905.02450} {Mass: Masked sequence to sequence pre-training for language generation}.

\bibitem[{Team(2022)}]{nllb2022}
NLLB Team. 2022.
\newblock No language left behind: Scaling human-centered machine translation.

\bibitem[{Vaswani et~al.(2017)Vaswani, Shazeer, Parmar, Uszkoreit, Jones, Gomez, Kaiser, and Polosukhin}]{vaswani2017attention}
Ashish Vaswani, Noam Shazeer, Niki Parmar, Jakob Uszkoreit, Llion Jones, Aidan~N. Gomez, Lukasz Kaiser, and Illia Polosukhin. 2017.
\newblock \href {http://arxiv.org/abs/1706.03762} {Attention is all you need}.

\bibitem[{Wang et~al.(2021)Wang, Yan, Meng, and Zhou}]{wang2021selective}
Fusheng Wang, Jianhao Yan, Fandong Meng, and Jie Zhou. 2021.
\newblock Selective knowledge distillation for neural machine translation.
\newblock In \emph{Proceedings of the 59th Annual Meeting of the Association for Computational Linguistics and the 11th International Joint Conference on Natural Language Processing (Volume 1: Long Papers)}, pages 6456--6466.

\bibitem[{Wang et~al.(2022)Wang, Jiao, Hao, Wang, Shi, Tu, and Lyu}]{wang-etal-2022-understanding}
Wenxuan Wang, Wenxiang Jiao, Yongchang Hao, Xing Wang, Shuming Shi, Zhaopeng Tu, and Michael Lyu. 2022.
\newblock \href {https://doi.org/10.18653/v1/2022.acl-long.185} {Understanding and improving sequence-to-sequence pretraining for neural machine translation}.
\newblock In \emph{Proceedings of the 60th Annual Meeting of the Association for Computational Linguistics (Volume 1: Long Papers)}, pages 2591--2600, Dublin, Ireland. Association for Computational Linguistics.

\bibitem[{Weng et~al.(2019)Weng, Yu, Huang, Cheng, and Luo}]{weng2019acquiring}
Rongxiang Weng, Heng Yu, Shujian Huang, Shanbo Cheng, and Weihua Luo. 2019.
\newblock \href {http://arxiv.org/abs/1912.01774} {Acquiring knowledge from pre-trained model to neural machine translation}.

\bibitem[{Wu et~al.(2020)Wu, Passban, Rezagholizade, and Liu}]{wu2020skip}
Yimeng Wu, Peyman Passban, Mehdi Rezagholizade, and Qun Liu. 2020.
\newblock \href {http://arxiv.org/abs/2010.03034} {Why skip if you can combine: A simple knowledge distillation technique for intermediate layers}.

\bibitem[{Yang et~al.(2022)Yang, Wang, Zhou, Zhao, Yu, Zhang, and Li}]{yang2022making}
Jiacheng Yang, Mingxuan Wang, Hao Zhou, Chengqi Zhao, Yong Yu, Weinan Zhang, and Lei Li. 2022.
\newblock \href {http://arxiv.org/abs/1908.05672} {Towards making the most of bert in neural machine translation}.

\bibitem[{Zhang et~al.(2018)Zhang, Xiang, Hospedales, and Lu}]{8578552}
Ying Zhang, Tao Xiang, Timothy~M. Hospedales, and Huchuan Lu. 2018.
\newblock \href {https://doi.org/10.1109/CVPR.2018.00454} {Deep mutual learning}.
\newblock In \emph{2018 IEEE/CVF Conference on Computer Vision and Pattern Recognition}, pages 4320--4328.

\bibitem[{Zhou et~al.(2021)Zhou, Neubig, and Gu}]{zhou2021understanding}
Chunting Zhou, Graham Neubig, and Jiatao Gu. 2021.
\newblock \href {http://arxiv.org/abs/1911.02727} {Understanding knowledge distillation in non-autoregressive machine translation}.

\bibitem[{Zhu et~al.(2020)Zhu, Xia, Wu, He, Qin, Zhou, Li, and Liu}]{zhu2020incorporating}
Jinhua Zhu, Yingce Xia, Lijun Wu, Di~He, Tao Qin, Wengang Zhou, Houqiang Li, and Tie-Yan Liu. 2020.
\newblock \href {http://arxiv.org/abs/2002.06823} {Incorporating bert into neural machine translation}.

\bibitem[{Zhu et~al.(2018)Zhu, Gong et~al.}]{lan2018knowledge}
Xiatian Zhu, Shaogang Gong, et~al. 2018.
\newblock Knowledge distillation by on-the-fly native ensemble.
\newblock \emph{Advances in neural information processing systems}, 31.

\end{thebibliography}
\newpage

\label{sec:appendix}

\end{document}